# "Even if" *Explanations:*
# *Prior Work, Desiderata & Benchmarks for Semi-Factual XAI*

Saugat Aryal[1,2] , Mark T. Keane[1,2]

[1] School of Computer Science, University College Dublin, Dublin, Ireland
[2] Insight Centre for Data Analytics, Dublin, Ireland
saugat.aryal@ucdconnect.ie, mark.keane@ucd.ie

## Abstract

Recently, eXplainable AI (XAI) research has focused on counterfactual explanations as post-hoc justifications for AI-system decisions (e.g., a customer refused a loan might be told "if you asked for a loan with a shorter term, it would have been approved"). Counterfactuals explain what changes to the input-features of an AI system change the output-decision. However, there is a sub-type of counterfactual, *semi-factuals*, that have received less attention in AI (though the Cognitive Sciences have studied them more). This paper surveys *semifactual explanation*, summarising historical and recent work. It defines key desiderata for semifactual XAI, reporting benchmark tests of historical algorithms (as well as a novel, naïve method) to provide a solid basis for future developments.

## 1  Introduction

With the emergence of deep learning there has been rising concern about the opacity of Artifical Intelligence (AI) systems and their impact on public and private life [Adadi and Berrada, 2018; Guidotti *et al.*, 2018]. Currently, governments are taking steps to protect people's rights, to regulate the AI industry and ensure that these technologies are not abused (e.g., the EU's GDPR [Goodman and Flaxman, 2017]). Research on eXplainable AI (XAI) tries to address such issues using automated explanations to improve the transparency of black-box models, to audit datasets and ensure fairness, accountability and trustworthiness [Gunning and Aha, 2019; Sokol and Flach, 2019; Birhane *et al.*, 2022].

Recently, significant research effort have been expended on counterfactual explanations for XAI [Byrne, 2019; Miller, 2019; Keane *et al.*, 2021; Karimi *et al.*, 2022]; a recent survey paper reports 350 papers on the topic [Verma *et al.*, 2022]. In this paper, we survey a less-researched special-case of the counterfactual, *semi-factual explanations*. In this review, we survey the literature on semi-factuals, we define desiderata for this strategy, identify key evaluation metrics and implement baselines to provide a solid base for future work.

Counterfactuals aim to explain algorithmic decisions in a post-hoc fashion, as an after-the-fact justification. So, in XAI, counterfactuals are typically used to explain what changes to the input-features of an AI system will *change* the outputdecision. For example, when a customer is refused a loan (i.e., the negative-class outcome for Q in Fig.1), the counterfactual might say "if you asked for a loan with a shorter term, it would have been approved" (the red C in Fig.1 that has a positive-class outcome). Technically, these could be called "outcome-counterfactuals" as they capture changes to the world that *change* the *outcome* (here, to be consistent with the literature, we will mostly call them "counterfactuals").

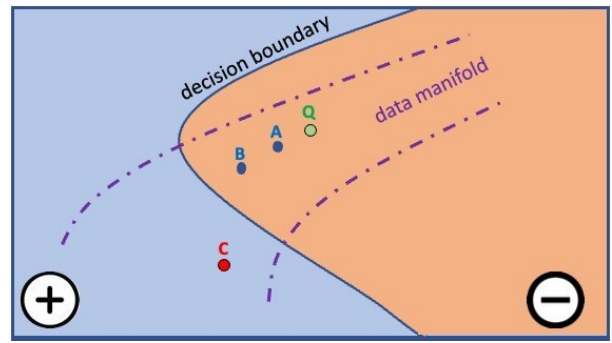

Figure 1: A and B are two semi-factuals (in blue) for the query Q (in green) all in the same class (i.e. the negative one), whereas the counterfactual C (in red) is over the decision boundary in the positive class. B is considered to be a better semi-factual than C, because B is further from Q and closer to the decision boundary.

Semi-factuals are a special-case of the counterfactual; they differ from outcome-counterfactuals in that they show endusers the feature changes that *do not change* a decisionoutcome. They are "counterfactual" in that they convey possibilities that "counter" what actually occurred, even though the outcome does *not* change. So, when the customer is refused the loan, the semi-factual might say "even if you doubled your stated income, you would still be refused the loan" or, indeed, "trebled your stated income" (the blue A and B in Fig.1 with the unchanged outcomes). Indeed, the larger the featuredifferences asserted in the semi-factual, the better (more convincing) the explanation (e.g., B is better than A in Fig.1). Philosophers have argued over whether semi-factuals really differ from outcome-counterfactuals (see [Bennett, 2003; Goodman, 1947]), but they have been shown to differ in their psychological impacts [McCloy and Byrne, 2002].

All the benefits accruing to counterfactuals in XAI also seem to accrue to semi-factuals; namely, that they have many legal [Wachter *et al.*, 2017], psychological [Byrne, 2019] and technical benefits [Keane *et al.*, 2021]. For example, in a medical domain, an intern reviewing X-rays of tumors could be told "even if this tumour was half its current size, it would still require a surgical intervention". Similarly, semi-factuals can convey key aspects of a casual model (e.g., a farmer might be told "even if you doubled your fertiliser use, your yield would not increase" because of soil factors). However, as we shall see, semi-factuals also differ significantly in many respects from counterfactuals (see desiderata, section 4).

Outline of Paper & Contributions: In this paper, we systematically review prior work on semi-factuals (henceforth, SFs) in the Cognitive Sciences and AI, beginning with a discussion of key examples from the early literature in Philosophy and Psychology (see section 2). From this work we define desiderata for SFs (section 3). In section 4, we report the results of a systematic survey before sketching the brief history of semi-factual algorithms for explanation (section 5). We then report a benchmarking study implementing key historical algorithms along with a newly-proposed na¨ive benchmark (see section 6), before closing with some conclusions (see section 7). As such, the paper makes several novel contributions to this emerging area of XAI, providing:

- A comprehensive survey of the relevant literature.

- A first statement of desiderata for semi-factual XAI.

- A na¨ive benchmark algorithm, based on the new idea of Most Distant Neighbors (MDNs).

- Novel comparative tests of historical benchmarks, to identify the best for future use.

- A publically-available repository of metrics, data, results for these benchmarks and an annotated bibliography (see *https://github.com/itsaugat/sf survey*).

---

¹ Because Philip is allergic to the ice-cream in both desserts.

## 2    Philosophy & Psychology of Semi-Factuals

Semi-factuals have been studied under different guises in Philosophy and Psychology for several decades. In Philosophy, counterfactuals (*if only...*) and semi-factuals (*even if...*) are often compared to conditionals (*if...then*) with a view to analysing their logic, truth conditions and role in causation [Chisholm, 1946; Goodman, 1947; Bennett, 1982; Barker, 1991; Bennett, 2003]. For example, [Bennett, 1982] and [Barker, 1991] argue about how the words "even" and "still" affect the interpretation of examples, such as:

(1) *Even if the United States had used nuclear weapons in Vietnam, it would still have lost the war.*

where the semi-factual asserts that *even if* the military-force expended by United States significantly increased, the Vietnam War would *still* have been lost. In AI terms, the semifactual says increasing the feature-value of *military-force* would not change the outcome. Hence, [Iten, 2002] proposes "scalar" analyses of *even* and *even if;* "Even Neville passed the exam" puts Neville low on an academic-ability scale.

In Psychology, semi-factual research has grown out of studies on counterfactual thinking in human cognition [Kahneman and Tversky, 1982; Byrne, 2007; Byrne, 2011; Handley and Feeney, 2007; Epstude and Roese, 2008]. Byrne [2007] proposed a mental model theory of semi-factuals that has been tested in several psychological studies (see e.g., [McCloy and Byrne, 2002; Parkinson and Byrne, 2017]). McCloy & Byrne's [2002] seminal work explicitly compared people's reasoning using matched scenarios for counterfactuals and semi-factuals, akin to the case of Philip who has an allergic reaction to an ice-cream sundae:

(2) *If only Philip had not chosen the ice-cream sundae, he wouldn't have had an allergic reaction.* (Counterfactual)

(3) *Even if Philip had chosen the banana split, he would still have had an allergic reaction*¹. (Semi-factual)

McCloy & Byrne found that counterfactuals lead people to judge the antecedent event (i.e., the choice of dessert) to be more causally-related to the outcome, but semi-factuals had the opposite effect, leading people to judge the antecedent event to be less causally-related to the outcome. So, semi-factuals weaken the causal link between the inputs and outcome, convincing people that outcome would have occurred anyway (people also differ in their emotional reactions to these events). In another experiment, they also found that counterfactuals lead people to focus on alternative antecedents that undo the outcome (e.g., "If only Philip had chosen the cheese cake he would not have had a reaction"), whereas semi-factuals lead people to focus on alternative antecedents that do *not undo* the outcome (e.g.,

"Even if Philip had chosen the baked-alaska he would still have had a reaction"). Subsequent studies test other psychological aspects of semi-factuals [Parkinson and Byrne, 2017; Moreno-Rios *et al.*, 2008; Santamaría *et al.*, 2005; Espino *et al.*, 2022].

Taken together these psychological findings show that semi-factuals have very different psychological effects than counterfactuals. Unlike counterfactuals, semi-factuals convince people of the *status quo*, they dissuade them from questioning outcomes [Green, 2008], and weaken the causal link between features and outcomes.

## 3   Desiderata for Semi-Factuals

Several desiderata are suggested by these analyses of semifactuals. These desiderata cover computational (i.e., "what needs to be computed") and psychological requirements (i.e., the response to be elicited in users) and are defined as follows. Assume (i) a query instance, $Q$, that has a vector, $x$, and an outcome, $y$, that occurs when $x$ holds and (ii) a semi-factual instance, $SF$, that has a vector, $x'$, and an outcome, $y'$, that occurs when $x'$ holds. $SF$ will be a good explanation of $Q$ if:

a) $Q$ is factually the case and $SF$ counters some of $Q's$ facts but not $Q's$ outcome; so the vectors $x$ and $x'$ differ, $diff(x, x')$, with no outcome change, $y = y'$.

b) Ideally, $SF$ relies on sparse changes to a key-feature(s), $f$, of $Q$, with other features being equal[2], ideally, one feature change (i.e., $diff(x,x')$=1); though many featuredifferences could be proposed, fewer is assumed to be better for psychological reasons.

c) The key-feature(s) changed should be plausible/mutable/actionable; that is, the $SF$ produced by the change should be within the data-manifold.

d) People should find the $SF$ convincing even though it may seem to be unexpected/surprising/counter-intuitive; for instance, they may expect the key-feature change to change the outcome, where $y/= y'$.

e) If people accept $SF$, it will change their perception of the causal role of the key-feature(s), $f$, in the domain. So, their causal model of the domain will change (e.g., causes may be updated/deleted/refined).

f) For fairness and ethical reasons, the asserted differences between $Q$ and $SF$, should not be misleading. For instance, (i) the key-feature should *not* be a proxy variable, (ii) the change asserted should be robust (e.g., resistant to adversarial attacks in that local region of the decision space), (iii) though the change may be unexpected it should not violate the domain's causality, (iii) the change assumes *ceteris paribus* (i.e., "other things being equal"), verifiably so (i.e., the unchanged-outcome shown should not depend on subtle interactions with other variables).

These desiderata present a high bar for semi-factual explanation methods; indeed, it is unclear whether any current method meets all of them. Furthermore, some of them may require further computational specification (e.g., how keyfeatures are selected) and psychological specification in operational definitions for user studies (e.g., for the notions of plausibility, convincingness and surprise).

## 4   Systematic Survey: *Even if* Explanations

A systematic search of the AI, Philosophy and Psychology literatures on semi-factuals was conducted using a bottomup citation-search and top-down keyword-searches (see Table 1). Ten searches were carried out between October $12^{th}$, 2022 and December $19^{th}$, 2022, consisting of (i) a bottomup search checking GoogleScholar citations to three key papers (i.e., [Cummins and Bridge, 2006; Nugent *et al.*, 2009; Kenny and Keane, 2021], (ii) nine top-down searches using keywords in GoogleScholar (see Table 1). The papers found (N=1,150) were title-and-abstract screened to check whether they were just citing semi-factuals or substantially researching them as a topic. Subsequent selections then identified the core papers of relevance (see *here* for PRISMA diagram).

| Search Terms | # | Papers Found | Unique Papers |
|---|---|---|---|
| *no search terms* (citation search of key papers) | 1 | 108 | 17 |
| "sf", "nearest-neighbor" | 2 | 20 | 3 |
| "sf", "ai" | 3 | 95 | 12 |
| "sf", "ai", "xp" | 4 | 86 | 12 |
| "sf", "xai" | 5 | 44 | 0 |
| "ai", "xp", ("near-hit" OR "nearest-hit") | 6 | 230 | 20 |
| "ai", "xp", "nearest-like neighbors" | 7 | 12 | 0 |
| "sf", "xp", "philosophy" | 8 | 203 | 11 |
| "sf", "xp", "psychology" | 9 | 228 | 3 |
| "xp", "even if conditionals", "linguistic", "philosophy" | 10 | 124 | 14 |
| *Totals* | | 1,150 | 92 |

Table 1: Ten searches used in the systematic survey of GoogleScholar (12-10-2022 to 19-12-2022) with the number of papers found and unique papers reviewed further (n.b., "sf", "ai" and "xp" are short for "semi-factual", "artificial intelligence" and "explanation", respectively).

---

[2] Equal may not mean the features have identical values, they may just be within some threshold difference.

## 4.1 Survey Results

Of the 1,150 original results checked, 92 potentially-relevant papers were selected to be read in depth from which 62 core papers were identified (41 cited here; note, 145 duplicates were removed). As we shall see in the next section on history (section 5), from a low base semi-factual research in AI has expanded considerably in the last two years. Note, many semi-factual papers in Philosophy, Psychology and Linguistics were checked but few are specifically relevant to explanation (e.g., in Philosophy the focus tends to be on the truth conditions of counterfactual statements and the linguistic functions of "even" and "still"). Finally, it should also be said that many excluded papers were from closely-related areas that do not cover semi-factuals *per se*, but which could provide insights for future work; areas that include research on (i) case difference learning (e.g., [Hanney and Keane, 1996; Ye *et al.*, 2021]), (ii) feature selection using near misses (e.g., [Kira *et al.*, 1992; Herchenbach *et al.*, 2022]), (iii) counterfactual explanation (e.g., [Keane *et al.*, 2021; Verma *et al.*, 2022]), (iv) flip-points in learning (e.g., [Yousefzadeh and O'Leary, 2019]), (v) dynamic critiquing in recommenders (e.g., [Reilly *et al.*, 2004]) and computational argumentation (e.g., [Cyras̆ *et al.*, 2021]. These papers are recorded in a publically-available annotated bibliography (see *https://github.com/itsaugat/sf survey*).

## 5 A Brief History of Semi-Factual XAI

In AI, semi-factuals have only been considered recently, relative to the philosophical and psychological literatures. Much of the initial work emerged from Case-Based Reasoning (CBR) research on post-hoc, example-based explanations [Sørmo *et al.*, 2005; Keane and Kenny, 2019]. In this AI research, semi-factual explanations have been variously cast as *a fortiori* arguments [Nugent *et al.*, 2005; Nugent *et al.*, 2009] and *precedent-based explanations* [Cummins and Bridge, 2006; Bridge and Cummins, 2005]. More recently, Kenny & Keane [2021] re-connected this work to the older literatures by calling them "semi-factuals". Arguably, there are four distinct phases in the development of semi-factual explanation in AI: (i) initial utility-based proposals, (ii) proximity-based methods, (iii) local-region methods and (iv) the more recent "modern-era" of counterfactuallyinspired proposals. In the following sub-sections, we describe each in turn and the intuitions behind them. We end this section by defining a new benchmark-method based on the notion of Most Distant Neighbors (MDNs).

## 5.1 Semi-Factuals Based on Feature-Utility

Doyle *et al.* [2004] appear as the first AI paper in our searches to propose using semi-factuals to explain automated decisions, under the rubric of *a fortiori reasoning*. An *a fortiori* argument is defined as one that uses a stronger version of an already-convincing proposition (i.e., "EU countries cannot afford standing armies, sure even the US can hardly afford its standing army"). Doyle *et al.* [2004] noted that nearestneighbor, example-based explanations can often be less convincing than neighbors that have more extreme feature-values within the same class. For example, if patient-x with a moderate temperature is judged to be dischargeable then a semifactual past case, patient-y with a much higher temperature who was discharged is more convincing than pointing to another patient with the same moderate temperature being discharged [Doyle *et al.*, 2006]. So, this semi-factual method computes a set of nearest neighbours as explanatory cases and then re-ranks them using utility functions on selected features to find a more convincing *a fortiori* case, as follows:

$$\text{Utility}(q,x,c) = \sum_{f \in F} w_f \xi \, (q_f, x_f, c) \qquad (1)$$

$$\text{SF}_{\text{Utility}}(q,x,c) = \arg \max_x Utility(q,x,c) \quad (2)$$

where $q$ is the query, $x$ is an instance, $c$ is a class label and $\xi()$ measures the contribution to explanation utility of the feature $f$. The $\xi()$ function uses relative-differences in feature-values to assign utilities. For example, for the temperature feature, the measure might assign higher utility to a $10^\circ$C difference than to a $5^\circ$C difference between a query and semi-factual case. This method priorities explanatory instances with more convincing feature-values, and may compute these over multiple features. Indeed, these utilities are seen as being classspecific and, even, user-specific, depending on what a given user may find convincing. Furthermore, [Doyle *et al.*, 2004] argued that these utility values often decreased as instances approach the decision boundary, as they were more likely to be outliers in the class and, therefore, less convincing. They user tested this method in a medical domain, showing that semi-factuals provided better explanations than the top-3 nearest neighbors for queries tested.

However, this method was knowledge-intensive, the utility values for each feature had to be hand-coded for each class (and, presumably, for each end-user). Indeed, in one of their user tests, the utility measures had to be re-defined half-way through the study to better reflect end-users' assessments [Doyle *et al.*, 2006]. This is a major drawback for the technique, as it begs the critical question about what featuredifferences will actually be more convincing. Accordingly, this utility method is not a plausible benchmark, though we do use their intuition about feature-differences to define a new, useful benchmark method (see section 5.4).

## 5.2 NUN-Related Semi-Factuals

Cummins & Bridge's [2006] "Knowledge-Light based Explanation-Oriented Retrieval" (KLEOR) approach proposed three methods based on similarity to Nearest Unlike Neighbors (NUNs). These KLEOR variants use the NUN to find the best semi-factual for a given query (n.b., they called the NUN, a *Nearest Miss*). In modern parlance, the NUN is the closest instance in the dataset bearing a counterfactual relationship to the query (see [Keane and Smyth, 2020]).

The first variant, *Sim-Miss*, selects an instance to be the semi-factual which is most similar to the NUN but in the same class as the query $q$:

$$\text{SF}_{\text{Sim-Miss}}(q,nun,G) = \underset{x \in G}{\text{argmax }} Sim(x,nun) \qquad (3)$$

where $q$ is the query, $x$ is the instance, $G$ represents the set of all instances in the same class as the query, and *nun* is the Nearest Unlike Neighbor, with *Sim* being Euclidean Distance or Cosine Similarity. This variant is the most naïve as it assumes a simple decision boundary. The second variant, *Global-Sim* method, is more sophisticated in that it requires the semi-factual be closer to $q$ than to the *nun* (to avoid SFs far from the query but close to the NUN):

$$\text{SF}_{\text{Global-Sim}}(q,nun,G) = \underset{x \in G}{\text{argmax }} Sim(x,nun)$$
$$+ Sim(q,x) > Sim(q,nun) \qquad (4)$$

using the global similarity between instances. Finally, the third variant, *Attr-Sim*, computes more fine-grained similarities for each feature-attribute, ensuring that the semi-factual lies between the $q$ and *nun* across the majority of features:

$$\text{SF}_{\text{Attr-Sim}}(q,nun,G) = \underset{x \in G}{\text{argmax}} Sim(x,nun)$$
$$+ \text{maxcount}[Sim(q_a,x_a) > Sim(q_a,nun_a)] \; a \epsilon F \qquad (5)$$

where $F$ is the feature-dimension set and $a$ is a featureattribute. These methods rely on the interesting intuition that a known counterfactual can be guide to finding a good semifactual explanation; that is, the semi-factual is to be found between the query and the NUN in the feature space.

Cummins & Bridge's carried out a computational evaluation of these three KLEOR variants using test-instances from a single dataset, showing that each method could find semifactuals for most queries. The less restrictive $SF_{Sim-Miss}$ method had the best coverage and, counter-intuitively, the most restrictive $SF_{Attr-Sim}$ did better than $SF_{Global-Sim}$. They also performed a psychological evaluation showing that $SF_{Sim-Miss}$ and $SF_{Attr-Sim}$ found semi-factuals that people thought to be as good as those found by $SF_{Utility}$, notably without the latter's knowledge engineering overheads. Accordingly, all three of these KLEOR variants were used in the present benchmarking study (see sect. 6).

## 5.3 Semi-Factuals & Local-Region Boundaries

Nugent *et al.* [2009] proposed another *a fortiori* method, by finding marginal instances in the local region around the query. Here, a surrogate model, specifically, logistic regression was used to capture the local neighborhood of the query, built using subset of instances located around it (akin to LIME [Ribeiro *et al.*, 2016]). This use of known instances ensures that the surrogate model learns about the local space surrounding the query, finding a reliable proxy justifying query's global behaviour. Finally, candidate nearest neighbors are tested using this local model to give a probability, with the marginal-probability instance closest to the decision boundary, being chosen as the semi-factual explanation, as follows:

$$\text{SF}_{\text{Local-Region}}(q,C) = \underset{x \in C}{\text{argmin}} LR(x) \qquad (6)$$

where, $C$ is the set of candidate neighbors and $LR()$ is the local logistic regression model providing the probability score.

The intuition here is that good semi-factuals should be close to the query's local decision boundary, while being as far as possible from it in this local space (as in Fig.1). So, a convincing semi-factual explanation should be locally close to the query but as distant from it as possible within this local region. Unfortunately, the authors did not computationally evaluate the method, though they did ask users to evaluate the explanations output. Their results showed that users had higher satisfaction and confidence in the semi-factual explanations compared to conventional nearest-neighbor examplebased explanations. Accordingly, this method is also used in the present benchmarking study (see section 6).

## 5.4 A New Benchmark: Most Distant Neighbors

Analogies between counterfactual XAI and semi-factuals suggest another naïve benchmark that has not been proposed before in the literature. Early counterfactual methods often used Nearest Unlike Neighbors (NUNs), the nearest classdifferent instance in the dataset to the query, as counterfactual explanations [Cunningham *et al.*, 2003; Wexler *et al.*, 2019]. NUNs can be a reasonable first-pass at counterfactuals that are guaranteed to be within-domain (though they have other weaknesses). An analogous solution for semi-factual explanations relies on the notion of Most Distant Neighbors (MDNs); namely, the *most distant same-class instance in the dataset to the query on some key-feature*. MDNs should be good semi-factuals because they reflect many of the desiderata and are, by definition, within-domain.

To compute MDNs, for a given feature of $q$, its neighbours on the dimension are partitioned into instance-sets that have higher values (i.e., HighSet) or lower values (i.e., LowSet) than the query. Each of these sets are ranked-ordered separately using a *Semi-Factual Scoring* (*sfs*) function, a distance messure that prioritises instances that are sparse (few feature differences) while also having the highest value-differences on a key-feature, as follows:

$$\text{sfs}(q,S,F) = \frac{same(q,x)}{F} + \frac{diff(q_f,x_f)}{diff_{max}(q_f,S_f)} \quad (7)$$

where $S$ is HigherSet or LowerSet and $x \in S$, $same()$ counts the features that are equal between $q$ and $x$, $F$ is the total number of features, $diff()$ gives the difference-value of keyfeature, $f$, and $diff_{max}()$ is the maximum difference-value for that key-feature in the HighSet/LowSet. Basically, the instance with the highest overall $sfs$ value from the HighSet/LowSet is the best candidate for that feature. This computation is done for each feature of $q$, independently, with the best of the best instances (i.e., with the highest $sfs$ value across all features) being chosen as the overall semifactual for the query (see Algorithm 1).

$$\text{SF}_{\text{MDN}}(q,S) = \underset{x \in S}{\text{argmax}}\ sfs(x)$$

The intuition behind MDNs is that if one can find an instance that has some features in common with the query but is as far from it on a key-feature, then it will make a good semifactual (see desiderata). This new method was also added to benchmarking study to compare it to the historical methods.

---

**Algorithm 1** *MDN Semi-factual*

**Input:** query $q$
**Output:** Semi-factual($q$)
1: Initialize $I = \emptyset,\ F = \emptyset,$
2: **for** feature $f = f_1, f_2, f_3, ..., f_n$ **do**
3:     $S = \{x : x_f \geq q_f \text{ or } x_f \leq q_f\}$    ▷ High/Low Set
4:     **for** $x \in S$ **do**
5:        $I \cup sfs(x)$        ▷ Equation 7
6:     **end for**
7:     $F \leftarrow F \cup \max(I)$
8: **end for**
9: $SF(q) \leftarrow \max(F)$
10: **return** $SF(q)$

---

### 5.5 The Modern Era: Post-2020 Methods

Kenny & Keane [2021] instigated, what could be called, the modern-era of semi-factual AI research with their GANbased counterfactual method, called PIECE, that also computed semi-factuals. PIECE finds "exceptional" and "normal" features for a given class and then modifies the query's "exceptional" features to create instances that have the "normal" features of the counterfactual class, using a GAN to generate image visualisations. As successive exceptionalfeatures are changed the generated instances move away from the query towards the counterfactual class, with the instance generated just before the decision boundary being chosen as the semi-factual. Kenny & Keane showed these semi-factuals to be more distant from the query than those found by other perturbation techniques (see Expt.2). In one sense, this solution re-imagines the Cummins-Bridge intuition that good semi-factuals can be found somewhere between the query and a counterfactual, close to the decision boundary.

PIECE kicked off a renewed interest in semi-factual XAI as researchers looked to improve it and to apply semi-factuals in different application contexts. So, Zhao *et al.* [2022] have proposed a class-to-class variational encoder (C2CVAR) which is less computationally expensive than PIECE that can generate semi-factuals (and counterfactuals; see also [Ye *et al.*, 2020]). Vats *et al.* [2022] have used StyleGAN2 [Karras *et al.*, 2020] to find semi-factual explanations for classifications of medical images of ulcers. Though these works try to explain *model capabilities*, others have proposed using semi-factuals to explain *model limits*. Artelt & Hammer [2022] use semi-factuals to explain the "reject option"; where an AI system rejects inputs because "a prediction with an unacceptable lower certainty" can only be made. Their perturbation-based optimisation method uses a loss function that promotes diverse semi-factuals that are (i) in the same class as the query (they are also rejected), (ii) sparse (they aim for 1-feature-difference), (iii) "sufficiently distant" from the query, and (iv) of higher certainty than the query (to make them more convincing). Notably, here, the key-feature being varied is the certainty of the instance's prediction. In a similar vein, Lu *et al.* [2022] argue that semi-factuals may be used to explain spurious patterns in human-in-the-loop ML (see also [Hagos *et al.*, 2022]). Finally, Mertes *et al.* [2022] propose an (apparently) wholly new type of counterfactual, called "alterfactuals", to explore the"irrelevant feature" space of the model; they describe these as semi-factuals that "move parallel to the decision boundary, indicating which features would not modify the model's decision".

Other similar proposals have also been made, though they suffer from a poor knowledge of the literature. [Fernandez´ *et al.*, 2022] propose a framework for contrastive explanations, called "explanation sets", in which they use a very loose, non-standard definition of semi-factuals (i.e., as basically any instance in some sub-region of the query's class), perhaps because they seem to be unaware of the prior literature. [Herchenbach *et al.*, 2022] also overlook prior work in their proposal of a broad framework involving *near hits* and near misses (counterfactuals) as image-explanations, though it is not clear whether the former are really semi-factuals (or just nearest neighbors).

Finally, from the user perspective, Mueller *et al.* [2021] include a semi-factual module in their cognitive tutorial for training users about "cognitively-challenging aspects of an AI system" and [Salimi, 2022] reports user-tests for trustworthiness after using semi-factuals.

These recent papers reflect a rapidly-expanding interest in semi-factual XAI. In time, these modern-era methods will need to be comparatively evaluated relative to the benchmarks and metrics proposed here, to determine which fare best in explaining predictions to end-users.

## 6 Benchmarking Study

To provide a firm empirical basis for future work on semifactual XAI, we ran a benchmark study of five methods, the four historical methods [i.e., the three KLEOR methods (SimMiss, Global-Sim, Attr-sim) and the Local-Region one] and the newly-proposed MDN method. Standard evaluation metrics from prior XAI work were used to compare these methods, using the five measures detailed below.

*Query-to-SF Distance*: The $L_2$-norm from the Query to the SF, where higher scores are better, as the semi-factual should be far from the query.

*Query-to-SF kNN (%)*: This is a measure of the percentage of instances (within the whole dataset) in the k-NN set surrounding the Query that occur before the SF is included (i.e., as $k$ is successively increased upto the appearance of the SF); it is an alternative measure for how far the SF is from the Query in the dataset, so higher values are better.

*SF-to-Query-Class Distance*: A within-distribution measure for the closeness of the SF to the distribution of the Query-Class using Mahalanobis distance [Chandra and others, 1936], where lower values mean the SF is closer to the query-class distribution.

*SF-to-NUN Distance*: The $L_2$-norm from the SF to the NUN, where lower scores are better as the semi-factual is closer to the class boundary.

*MDN Distance*: The *sfs* function, a semi-factual-oriented distance from a Query to a SF, can used to determine how far the SFs selected by historical methods are from the Query; this metric allows us to assess whether historical methods find "better" MDNs than the MDN-method itself, where higher *sfs* values indicate the SF is a better MDN for the Query.

*Sparsity (%)*: The $L_0$-norm counting the number of feature-differences between the Query and SF, divided into three levels (i.e., 1-diff, 2-diff and >3-diff) where the percent of SFs selected by the method at each level is recorded; obviously, methods with higher percentage at lower difference levels are better (ideally, high-percentage at the 1-diff level).

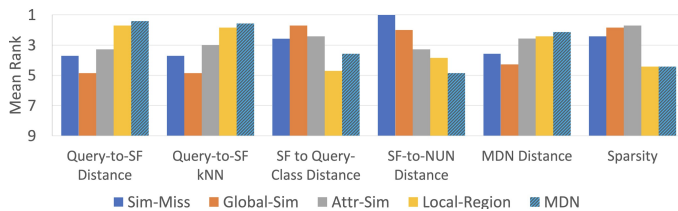

Figure 2: Mean Ranks of Success for the Five Benchmark Methods on Six Different Measures, over the Datasets Tested.

### 6.1 Method

We performed leave-one-out cross-validation for each of the five methods on seven datasets to find a semi-factual for every instance in the dataset, treating each as a query. We used 3-NN model to implement the KLEOR variants. For the Local Region method, we consider a minimum of 200 instances from each class to build the local model for a query. In the MDN method, a "20% of the standard deviation" threshold was used to determine whether values for a given feature were essentially "the same". The seven datasets were benchmark, publically-available, tabular datasets commonly used in the counterfactual literature, all binary-classed: Adult-Income (N=26,540, 12 features), Blood Alcohol (N=2,000, 5 features), Default Credit Card (N=30,000, 23 features), Pima Diabetes (N=392, 8 features), German Credit (N=1,000, 20 features), HELOC (8,291 instances, 20 features), Lending Club (N=39,239, 8 features). All the experiments were carried out in Python 3.9 on Ubuntu 16.04 machine with 40 core Intel Xeon(R) processor with an approximate run-time of 40 hours. All programs, data and results are available at *https://github.com/itsaugat/sf_survey*.

### 6.2 Results & Discussion

Figure 2 summarises the overall results for the five methods (as mean ranks over datasets) on the six benchmark measures (Figures 3, 4 and 5 show results by-dataset). The summary shows that MDN does best on three of the six measures (i.e., Query-to-SF Distance, Query-to-SF kNN, MDN Distance), with the Local Region method being a close second; performance on the other metrics (SF-to-Query-Class Distance, SFto-NUN Distance and Sparsity) requires some interpretation.

On the *Query-to-SF Distance* metric (Figure 3a) it can be seen that MDN produces the highest *Query-to-SF distances* for 4 of the 7 datasets, showing that it tends to find the furthest SF-instances from the query. On the *Query-SF kNN* metric (Figure 3b) MDN again scores the highest in 3 of 7 datasets with overall percentages that stand out; so, MDN finds SFs separated from the Query by many instances. On the *MDNDistance* metric (Figure 4) the four historical methods mainly produce lower scores across datasets (except for the HELOC dataset) showing that the MDN method is finding the furthest SFs from the Query in the dataset (i.e., it is finding the best MDNs in each dataset).

With respect to the other metrics MDN does less well. On the *SF-to-Q-Class Distance* measure (Figure 3c) MDN is ranked 4th; though all its SFs are by-definition within distribution (as valid datapoints), MDN probably scores lower as it is finding outliers in the distribution. On the *SF-to-NUN Distance* metric (Figure 3d), the KLEOR variants perform relatively better than the MDN and Local Region methods; however, this result is not that surprising as these methods are optimised to find SFs close to NUNs. Finally, on the *sparsity* measure (see Figure 5) MDN markedly differs from all other methods. In Figure 5, the higher the blue-portion of the bars [i.e., the % of 1-diff SFs] for a given method-dataset pair, the better the performance. MDN has very little blue, it is the worst of all the methods in three datasets where 100% of its SFs have >3-feature-differences (though in three others it fares better). This performance could probably be improved by fine-tuning the *sfs* function [see formula (7)]. Recall, that this function has two equally-weighted components, that compute (i) same-features and (ii) relative-differences in the key-feature. If a higher weight was given to the same-features component, then the method should select sparser SFs (perhaps also aided by a scoring threshold). For the present work, we felt it was better to provide a vanilla *sfs* function to get a clear sense of how a baseline-MDN method might work.

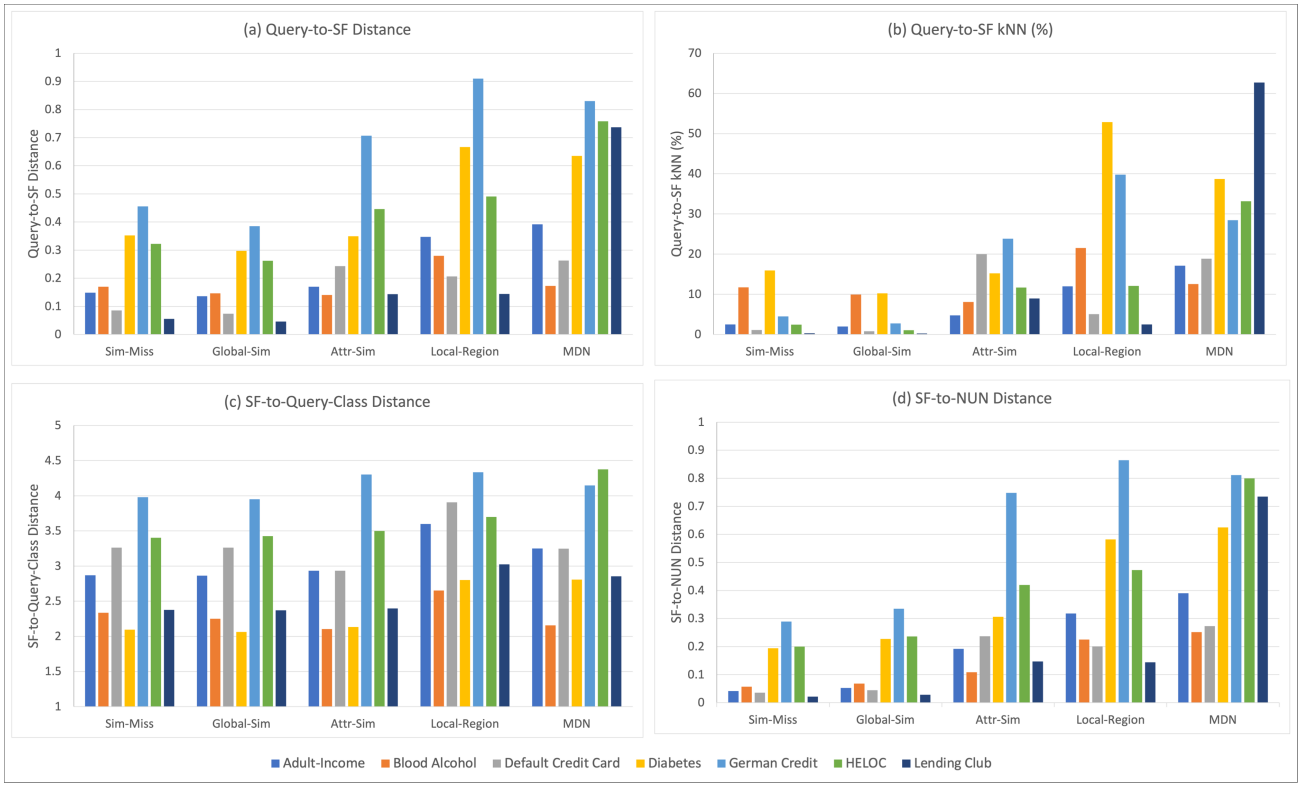

Figure 3: Benchmark Results: Performance of Five Semi-Factual Methods on Seven Tabular Datasets for Four Key Evaluation Measures, the (a) Query-to-SF Distance, (b) Query-to-SF kNN (%), (c) SF-to-Q-Class Distance, (d) SF-to-NUN Distance Measures.

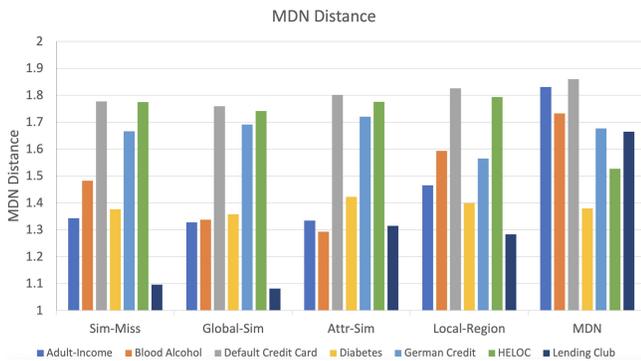

Figure 4: Results of MDN Distance Measure for the Semi-Factual Methods across Seven Datasets.

Overall, in conclusion, though it seems that the MDN and the Local Region methods provide the best candidates for semi-factual baselines. The Local Region method provides reasonable, solid results with decent sparsity, whereas the MDN method shows the furthest SF-point in the dataset from the Query (as type of upper limit to beat).

## 7 Conclusion

In recent years, counterfactual explanations have been heavily researched as a significant explanation strategy in XAI. Yet, very little attention has been given to an, arguably, equally useful method that relies on semi-factuals (where changes to input features do *not* lead to output changes). In this paper, from a systematic survey, we aim to remedy this deficit and place this topic area on a firm footing with defined desiderata, benchmarked methods and suitable metrics.

In conclusion, several limitations and caveats are to be noted.

With respect to limitations, it is to be noted that in the current benchmark study we have concentrated on tabular data, largely to respect the focus of historical methods. However, the desiderata and evaluation metrics should equally apply to image dataset (and possibly time-series data), albeit relying more on latent features (as has been demonstrated in [Kenny and Keane, 2021]). The paucity of user studies is another severe limitation; until some carefully-controlled studies are carried out, we do not really know how users will respond to these explanations in the AI context.

With respect to caveats, we believe that it is important to reiterate the ethical point about the use of semi-factuals (a point that also applies to counterfactuals [Asher *et al.*, 2022]). These explanatory methods have significant cognitive impacts on people's understanding of AI systems and domains, they convince and dissuade people. But, they could be misused if certain assumptions are violated (e.g., if the SF is not robust). So, future implementations of these methods will need to provide metrics to audit these assumptions, to ensure they are being properly and fairly applied in advice to end-users.

## Acknowledgments

This publication has emerged from research conducted with the financial support of Science Foundation Ireland (SFI) to the *Insight Centre for Data Analytics* under Grant Number 12/RC/2289 P2.

# Annotated Bibliography

\*    means cited in paper
SF_AI    means core papers related to SFs in AI
SF_PSY    means articles related to SFs in Psychology
SF_PHL    means papers related to SFs in Philosophy
CF    means related to Counterfactual XAI
SURV    means survey/review article related to XAI
REL    means areas closely related to SF

\*SURV [Adadi and Berrada, 2018] Amina Adadi and Mohammed Berrada. Peeking inside the black-box: A survey on explainable artificial intelligence (xai). *IEEE Access*, 6:52138–52160, 2018.

SF_AI [Armengol and Plaza, 2006] Eva Armengol and Enric Plaza. Symbolic explanation of similarities in case-based reasoning. *Computing and informatics*, 25(2-3):153–171, 2006.

\*SF_AI [Artelt and Hammer, 2022] André Artelt and Barbara Hammer. " even if..."–diverse semifactual explanations of reject. *arXiv preprint arXiv:2207.01898*, 2022.

\*CF [Asher et al., 2022] Nicholas Asher, Lucas De Lara, Soumya Paul, and Chris Russell. Counterfactual models for fair and adequate explanations. *Machine Learning and Knowledge Extraction*, 4(2):316–349, 2022.

\*SF_PHL [Barker, 1991] Stephen Barker. " even, still" and counterfactuals. *Linguistics and Philosophy*, pages 1–38, 1991.

SF_PHL [Barker, 1994] Stephen J Barker. The consequententailment problem foreven if. *Linguistics and Philosophy*, 17(3):249–260, 1994.

\*SF_PHL [Bennett, 1982] Jonathan Bennett. Even if. *Linguistics and Philosophy*, 5(3):403–418, 1982.

\*SF_PHL [Bennett, 2003] Jonathan Bennett. *A philosophical guide to conditionals*. Clarendon Press, 2003.

\*REL [Birhane et al., 2022] Abeba Birhane, Vinay Uday Prabhu, and John Whaley. Auditing saliency cropping algorithms. In *Proceedings of the IEEE/CVF Winter Conference on Applications of Computer Vision*, pages 4051–4059, 2022.

REL [Bolon-Canedo and Remeseiro, 2020] Veronica BolonCanedo and Beatriz Remeseiro. Feature selection in image analysis: a survey. *Artificial Intelligence Review*, 53(4):2905–2931, 2020.

REL [Booth et al., 2021] Serena Booth, Yilun Zhou, Ankit Shah, and Julie Shah. Bayes-trex: a bayesian sampling approach to model transparency by example. In *Proceedings of the AAAI Conference on Artificial Intelligence*, volume 35, pages 11423–11432, 2021.

SF_PHL [Booth, 2014] Charles Booth. *Boundary work in theory and practice: Past, present and future*. PhD thesis, University of the West of England, 2014.

SF_PSY [Branscombe et al., 1996] Nyla R Branscombe, Susan Owen, Teri A Garstka, and Jason Coleman. Rape and accident counterfactuals: Who might have done otherwise and would it have changed the outcome? 1.

SF_PHL [Butcher, 1983] David Butcher. An incompatible pair of subjunctive conditional modal axioms. *Philosophical Studies: An International Journal for Philosophy in the Analytic Tradition*, 44(1):71–110, 1983.

SF_PSY [Byrne, ] Ruth MJ Byrne. Counterfactuals, causes and exceptions.

SF_PSY [Byrne, 2007a] Ruth MJ Byrne. Precis of the rational imagination: How people create alternatives to reality. *Behavioral and Brain Sciences*, 30(5-6):439– 453, 2007.

\*SF_PSY [Byrne, 2007b] Ruth MJ Byrne. *The rational imagination: How people create alternatives to reality*. MIT press, 2007.

\*CF [Byrne, 2019] Ruth MJ Byrne. Counterfactuals in explainable artificial intelligence (xai): evidence from human reasoning. In *Proceedings of the Twenty-Eighth International Joint Conference on Artificial Intelligence, IJCAI- 19*, pages 6276–6282, 2019.

REL [Carvalho, 2022] Maria Manuel Domingos Carvalho. Towards biometrically-morphed medical case-based explanations. 2022.

\*SF_PHL [Chisholm, 1946] Roderick M Chisholm. The contrary-to-fact conditional. *Mind*, 55(220):289–307, 1946.

CF [Cho and Shin, 2023] Soo Hyun Cho and Kyung-shik Shin. Feature-weighted counterfactual-based explanation for bankruptcy prediction. *Expert Systems with Applications*, 216:119390, 2023.

REL [Craw et al., 2006] Susan Craw, Nirmalie Wiratunga, and Ray C Rowe. Learning adaptation knowledge to improve case-based reasoning. *Artificial intelligence*, 170(16- 17):1175–1192, 2006.

\*SF_AI [Cummins and Bridge, 2006] Lisa Cummins and Derek Bridge. Kleor: A knowledge lite approach to explanation oriented retrieval. *Computing and Informatics*, 25(2- 3):173–193, 2006.

\*SF_AI [Cunningham et al., 2003] Pádraig Cunningham, Dónal Doyle, and John Loughrey. An evaluation of the usefulness of case-based explanation. In *International conference on case-based reasoning*, pages 122–130. Springer, 2003.

CF [Dandl et al., 2020] Susanne Dandl, Christoph Molnar, Martin Binder, and Bernd Bischl. Multi-objective counterfactual explanations. In *International Conference on Parallel Problem Solving from Nature*, pages 448–469. Springer, 2020.

REL [Dash and Liu, 1997] Manoranjan Dash and Huan Liu. Feature selection for classification. *Intelligent data analysis*, 1(1-4):131–156, 1997.

SF_PHL [Declerck and Reed, 2001] Renaat Declerck and Susan Reed. Some truths and nontruths about even if. *Linguistics*, 39:203–255, 01 2001.

CF [Dhurandhar et al., 2018] Amit Dhurandhar, Pin-Yu Chen, Ronny Luss, Chun-Chen Tu, Paishun Ting, Karthikeyan Shanmugam, and Payel Das. Explanations based on the missing: Towards contrastive explanations with pertinent negatives. *Advances in neural information processing system*s, 31, 2018.

\*SF_AI [Doyle et al., 2004] Dónal Doyle, Pádraig

*Reasoning Research and Development: 30th International Conference, ICCBR 2022, Nancy, France, September 12–15, 2022, Proceedings*, pages 289–303. Springer, 2022.

*SF_PSY [Epstude and Roese, 2008] Kai Epstude and Neal J Roese. The functional theory of counterfactual thinking. *Personality and Social Psychology Review*, 12(2):168–192, 5 2008.

*SF_PSY [Espino *et al.*, 2022] Orlando Espino, Isabel Orenes, and Sergio Moreno-R´ıos. Inferences from the negation of counterfactual and semifactual conditionals. *Memory & Cognition*, 50(5):1090–1102, 2022.

SF_PSY [Feeney *et al.*, 2011] Aidan Feeney, Simon J Handley, et al. Suppositions, conditionals, and causal claims. *Under- standing counterfactuals and causation: Issues in philosophy and psychology*, pages 242–262, 2011.

CF [Feiman, 2008] Roman Feiman. Possible worlds and counterfactuals: Critique and commentary on complicating causation. *Episteme*, 19(1):4, 2008.

*SF_AI [Ferna´ndez *et al.*, 2022] Rube´n R Ferna´ndez, Isaac Mart´ın de Diego, Javier M Moguerza, and Francisco Herrera. Explanation sets: A general framework for machine learning explainability. *Information Sciences*, 617:464–481, 2022.

REL [Freitas *et al.*, 2008] Alex A Freitas, Daniela C Wieser, and Rolf Apweiler. On the importance of comprehensible classification models for protein function prediction. *IEEE/ACM Transactions on Computational Biology and Bioinformatics*, 7(1):172–182, 2008.

SURV [Gates and Leake, 2021] Lawrence Gates and David Leake. Evaluating cbr explanation capabilities: Survey and next steps. In *ICCBR Workshops*, pages 40–51, 2021.

SF_PHL [Gomes, 2020] Gilberto Gomes. Concessive conditionals without even if and nonconcessive conditionals with even if. *Acta Analytica*, 35(1):1–21, 2020.

REL [Goodman and Flaxman, 2017] Bryce Goodman and Seth Flaxman. European union regulations on algorithmic decision-making and a "right to explanation". *AI magazine*, 38(3):50–57, 2017.

*SF_PHL [Goodman, 1947] Nelson Goodman. The problem of counterfactual conditionals. *The Journal of Philosophy*, 44(5):113–128, 1947.

*SF_PSY [Green, 2008] David W Green. Persuasion and the contexts of dissuasion: Causal models and informal arguments. *Thinking & reasoning*, 14(1):28–59, 2008.

*SURV [Guidotti *et al.*, 2018] Riccardo Guidotti, Anna Monreale, Salvatore Ruggieri, Franco Turini, Fosca Giannotti, and Dino Pedreschi. A survey of methods for explaining black box models. *ACM computing surveys (CSUR)*, 51(5):1–42, 2018.

SF_PHL [Gu¨ngo¨r, ] Hu¨seyin Gu¨ngo¨r. Truthmaking even ifs.

*SURV [Gunning and Aha, 2019] David Gunning and David W Aha. Darpa's explainable artificial intelligence program. *AI Magazine*, 40(2):44-58, 2019.

*SF_AI [Hagos *et al.*, ] Misgina Tsighe Hagos, Kathleen M Curran, and Brian Mac Namee. Identifying spurious correlations and correcting them with an casebase. In *European Workshop on Advances in Case-Based Reasoning*, pages 179–192. Springer, 1996.

*SF_AI [Herchenbach *et al.*, 2022] Marvin Herchenbach, Dennis Mu¨ller, Stephan Scheele, and Ute Schmid.

Explaining image classifications with near misses, near hits and prototypes. In *International Conference on Pattern Recognition and Artificial Intelligence*, pages 419–430. Springer, 2022.

CF [Ho¨ltgen *et al.*, 2021] Benedikt Ho¨ltgen, Lisa Schut, Jan M Brauner, and Yarin Gal. Deduce: Generating counterfactual explanations efficiently. *arXiv preprint arXiv:2111.15639*, 2021.

*SF_PHL [Iten, 2002] Corinne Iten. Even if and even: The case for an inferential scalar account. *UCL Working Papers in Linguistics*, 14:119, 2002.

REL [Jalali *et al.*, 2017] Vahid Jalali, David Leake, and Najmeh Forouzandehmehr. Learning and applying case adaptation rules for classification: An ensemble approach. In *IJCAI*, pages 4874–4878, 2017.

*SF_PSY [Kahneman and Tversky, 1982] Daniel Kahneman and Amos Tversky. The Simulation Heuristic. In Daniel Kahneman, Paul Slovic, and Amos Tversky, editors, *Judgment Under Uncertainty: Heuristics and Biases*, pages 201–8. Cambridge University Press, New York, 1982.

*SURV [Karimi *et al.*, 2022] Amir-Hossein Karimi, Gilles Barthe, Bernhard Scho¨lkopf, and Isabel Valera. A survey of algorithmic recourse: contrastive explanations and consequential recommendations. *ACM Computing Surveys*, 55(5):1– 29, 2022.

*REL [Karras *et al.*, 2020] Tero Karras, Samuli Laine, Miika Aittala, Janne Hellsten, Jaakko Lehtinen, and Timo Aila. Analyzing and improving the image quality of stylegan. In *Proceedings of the IEEE/CVF conference on computer vision and pattern recognition*, pages 8110–8119, 2020.

*SURV [Keane and Kenny, 2019] Mark T Keane and Eoin M Kenny. How case-based reasoning explains neural networks: A theoretical analysis of xai using post-hoc explanation-by-example from a survey of ann-cbr twinsystems. In *International Conference on Case-Based Reasoning*, pages 155–171. Springer, 2019.

*CF [Keane and Smyth, 2020] Mark T Keane and Barry Smyth. Good counterfactuals and where to find them: A case- based technique for generating counterfactuals for explainable ai (xai). In *Proceedings of the 28th International Conference on Case-Based Reasoning (ICCBR-20)*, pages 163–178. Springer, 2020.

*CF [Keane *et al.*, 2021] Mark T Keane, Eoin M Kenny, Eoin Delaney, and Barry Smyth. If only we had better counter-factual explanations. In *Proceedings of the 30th International Joint Conference on Artificial Intelligence (IJCAI-21)*, 2021.

*SF_AI [Kenny and Keane, 2021] Eoin M. Kenny and Mark T. Keane. On generating plausible counterfactual and semi-factual explanations for deep learning. In *Proceedings of the 35th AAAI Conference on Artificial Intelligence (AAAI-21)*, pages 11575–11585, 2021.

*REL [Kira *et al.*, 1992] Kenji Kira, Larry A Rendell, et al. The feature selection problem: Traditional methods and a new algorithm. In *Aaai*, volume 2, pages 129–134, 1992.

REL [Liao *et al.*, 2018] Chieh-Kang Liao, Alan Liu, and Yu-Sheng Chao. A machine learning approach to case adaptation. In *2018 IEEE First International Conference on Artificial Intelligence and Knowledge Engineering (AIKE)*, pages 106–109. IEEE, 2018.

REL [Lin and Shaw, 1997] Fu-Ren Lin and Michael J Shaw. Active training of backpropagation neural networks using the learning by experimentation methodology. *Annals of Operations Research*, 75:105–122, 1997.

REL [Lou and Obradovic, 2012] Qiang Lou and Zoran Obradovic. Margin-based feature selection in incomplete data. In *Proceedings of the AAAI Conference on Artificial Intelligence*, volume 26, pages 1040–1046, 2012.

*SF_AI [Lu *et al.*, 2022] Jinghui Lu, Linyi Yang, Brian Mac Namee, and Yue Zhang. A rationale-centric framework for human-in-the-loop machine learning. *arXiv preprint arXiv:2203.12918*, 2022.

REL [Luss *et al.*, 2021] Ronny Luss, Pin-Yu Chen, Amit Dhurandhar, Prasanna Sattigeri, Yunfeng Zhang, Karthikeyan Shanmugam, and Chun-Chen Tu. Leveraging latent features for local explanations. In *Proceedings of the 27th ACM SIGKDD Conference on Knowledge Discovery & Data Mining*, pages 1139–1149, 2021.

SF_PHL [Lycan, 1991] William G Lycan. Even" and" even if. *Linguistics and Philosophy*, pages 115–150, 1991.

REL [McCarthy *et al.*, 2005] Kevin McCarthy, James Reilly, Lorraine McGinty, and Barry Smyth. Experiments in dynamic critiquing. In *Proceedings of the 10th international conference on Intelligent user interfaces*, pages 175–182, 2005.

*SF_PSY [McCloy and Byrne, 2002] Rachel McCloy and Ruth MJ Byrne. Semifactual "even if" thinking. *Thinking & Reasoning*, 8(1):41–67, 2002.

SF_AI [McSherry, 2004] David McSherry. Explaining the pros and cons of conclusions in cbr. In *European Conference on Case-Based Reasoning*, pages 317–330. Springer, 2004.

*SF_AI [Mertes *et al.*, 2022] Silvan Mertes, Christina Karle, Tobias Huber, Katharina Weitz, Ruben Schlagowski, and Elisabeth Andre´. Alterfactual explanations–the relevance of irrelevance for explaining ai systems. *arXiv preprint arXiv:2207.09374*, 2022.

SF_PHL [Kvart, 2001] Igal Kvart. The counterfactual analysis of cause. *Synthese*, 127(3):389–427, 2001.

CF [Laugel *et al.*, 2019] Thibault Laugel, Marie-Jeanne Lesot, Christophe Marsala, Xavier Renard, and Marcin Detyniecki. The dangers of post-hoc interpretability: Unjustified counterfactual explanations. *arXiv preprint arXiv:1907.09294*, 2019.

REL [Leake *et al.*, 2022] David Leake, Zachary Wilkerson, and David Crandall. Extracting case indices from convolutional neural networks: A comparative study. In *Case-Based Reasoning Research and Development: 30th International Conference, ICCBR 2022, Nancy, France, September 12–15, 2022, Proceedings*, pages 81–95. Springer, 2022.

*SURV [Miller, 2019] Tim Miller. Explanation in artificial intelligence: Insights from the social sciences. *Artificial Intelligence*, 267:1–38, 2019.

REL [Montenegro *et al.*, 2021] Helena Montenegro, Wilson

counterfactual explanations. In *Proceedings of the 2020 conference on fairness, accountability, and transparency*, pages 607-617, 2020.

*SF_AI [Mueller *et al.*, 2021] Shane Mueller, Yin-Yin Tan, Anne Linja, Gary Klein, and Robert Hoffman. Authoring guide for cognitive tutorials for artificial intelligence: Purposes and methods. 2021.

REL [Nimmy *et al.*, 2023] Sonia Farhana Nimmy, Omar K Hussain, Ripon K Chakrabortty, Farookh Khadeer Hussain, and Morteza Saberi. Interpreting the antecedents of a predicted output by capturing the interdependencies among the system features and their evolution over time. *Engineering Applications of Artificial Intelligence*, 117:105596, 2023.

*SF_AI [Nugent *et al.*, 2005] Conor Nugent, Pa´draig Cunningham, and Do´nal Doyle. The best way to instil confidence is by being right. In *International Conference on Case-Based Reasoning*, pages 368–381. Springer, 2005.

*SF_AI [Nugent *et al.*, 2009] Conor Nugent, Do´nal Doyle, and Pa´draig Cunningham. Gaining insight through casebased explanation. *Journal of Intelligent Information Systems*, 32(3):267–295, 2009.

SF_AI [Olsson *et al.*, 2014] Tomas Olsson, Daniel Gillblad, Peter Funk, and Ning Xiong. Explaining probabilistic fault diagnosis and classification using case-based reasoning. In *International Conference on Case-Based Reasoning*, pages 360–374. Springer, 2014.

REL [Olsson *et al.*, 2014] Tomas Olsson, Daniel Gillblad, Peter Funk, and Ning Xiong. Case-based reasoning for explaining probabilistic machine learning. *International Journal of Computer Science and Information Technology*, 6:87–101, 2014.

*SF_PSY [Parkinson and Byrne, 2017] Mary Parkinson and Ruth MJ Byrne. Counterfactual and semi-factual thoughts in moral judgements about failed attempts to harm. *Thinking & Reasoning*, 23(4):409–448, 2017.

SF_PHL [Paul, 2009] Laurie Ann Paul. Counterfactual theories. *The Oxford handbook of causation*, pages 158–184, 2009.

CF [Pedapati *et al.*, 2020] Tejaswini Pedapati, Avinash Balakrishnan, Karthikeyan Shanmugam, and Amit Dhurandhar. Learning global transparent models consistent with local contrastive explanations. *Advances in neural information processing systems*, 33:3592–3602, 2020.

SF_AI [Rabold *et al.*, 2022] Johannes Rabold, Michael Siebers, and Ute Schmid. Generating contrastive explanations for inductive logic programming based on a near miss approach. *Machine Learning*, 111(5):1799–1820, 2022.

REL [Raman and Ioerger, 2003] Baranidharan Raman and Thomas R Ioerger. Enhancing learning using feature and example selection. *Journal of Machine Learning Research (submitted for publication)*, 2003.

*REL [Reilly *et al.*, 2004] James Reilly, Kevin McCarthy, Lorraine McGinty, and Barry Smyth. Dynamic critiquing. In *European Conference on Case-Based Reasoning*, pages 763–777. Springer, 2004.

*REL [Ribeiro *et al.*, 2016] Marco Tulio Ribeiro, Sameer Singh, and Carlos Guestrin. " why should i trust you?" explaining the predictions of any classifier. In *Proceedings of the 22nd ACM SIGKDD international conference on*

REL [Rissland and Skalak, 1989b] Edwina L Rissland and David B Skalak. Interpreting statutory predicates. In *Proceedings of the 2nd international conference on Artificial intelligence and law*, pages 46–53, 1989.

SURV [Rissland, 2006] Edwina L Rissland. Ai and similarity. *IEEE Intelligent Systems*, 21(03):39–49, 2006.

SURV [Rissland, 2009] Edwina L Rissland. Black swans, gray cygnets and other rare birds. In *International Conference on Case-Based Reasoning*, pages 6–13. Springer, 2009.

*SF_AI [Salimi, 2022] Pedram Salimi. Addressing trust and mutability issues in xai utilising case based reasoning. *ICCBR Doctoral Consortium 2022*, 1613:0073, 2022.

SF_PSY[Santamar´ıa *et al.*, 2005] Carlos Santamar´ıa, Orlando Espino, and Ruth MJ Byrne. Counterfactual and semifactual conditionals prime alternative possibilities. *Journal of Experimental Psychology: Learning, Memory, and Cognition*, 31(5):1149, 2005.

SF_PHL [Sarasvathy, 2021] Saras D Sarasvathy. Even-if: Sufficient, yet unnecessary conditions for worldmaking. *Organization Theory*, 2(2):2631787721005785, 2021.

CF [Schleich *et al.*, 2021] Maximilian Schleich, Zixuan Geng, Yihong Zhang, and Dan Suciu. Geco: quality counterfactual explanations in real time. *arXiv preprint arXiv:2101.01292*, 2021.

*CF [Sokol and Flach, 2019] Kacper Sokol and Peter A Flach. Counterfactual explanations of machine learning predictions: opportunities and challenges for ai safety. *SafeAI@ AAAI*, 2019.

*SURV [Sørmo *et al.*, 2005] Frode Sørmo, Jo¨rg Cassens, and Agnar Aamodt. Explanation in case-based reasoning– perspectives and goals. *Artificial Intelligence Review*, 24(2):109–143, 2005.

REL [Sun, 2007] Yijun Sun. Iterative relief for feature weighting: algorithms, theories, and applications. *IEEE transactions on pattern analysis and machine intelligence*, 29(6):1035– 1051, 2007.

SF_PHL [Tellings, 2017] Jos Tellings. Still as an additive particle in conditionals. In *Semantics and Linguistic Theory*, vol- ume 27, pages 1–21, 2017.

REL [Urbanowicz *et al.*, 2018] Ryan J Urbanowicz, Melissa Meeker, William La Cava, Randal S Olson, and Jason H Moore. Relief-based feature selection: Introduction and review. *Journal of biomedical informatics*, 85:189–203, 2018.

*SF_AI [Vats et al., 2022] Anuja Vats, Ahmed Mohammed, Marius Pedersen, and Nirmalie Wiratunga. This changes to that: Combining causal and non-causal explanations to generate disease progression in capsule endoscopy. arXiv preprint arXiv:2212.02506, 2022.

*SURV [Verma *et al.*, 2022] Sahil Verma, Varich Boonsanong, Minh Hoang, Keegan E. Hines, John P. Dickerson, and Chirag Shah. Counterfactual explanations and algorithmic recourses for machine learning: A review, 2022.

SF_PSY [Vidal and Baratgin, 2017] Mathieu Vidal and Jean Baratgin. A psychological study of unconnected conditionals. *Journal of Cognitive Psychology*, 29(6):769–781, 2017.

SF_PHL [Vidal, 2017] Mathieu Vidal. A compositional semantics for 'even if' conditionals. *Logic and Logical*

Palihawadana, David Corsar, and Kyle Martin. How close is too close? the role of feature attributions in discovering counterfactual explanations. In *Case-Based Reasoning Research and Development: 30th International Conference, ICCBR 2022, Nancy, France, September 12–15, 2022, Proceedings*, pages 33–47. Springer, 2022.

SF_AI [Winston, 1970] Patrick H Winston. Learning structural descriptions from examples. 1970.

CF [Wiratunga *et al.*, 2021] Nirmalie Wiratunga, Anjana Wi- jekoon, Ikechukwu Nkisi-Orji, Kyle Martin, Chamath Pal- ihawadana, and David Corsar. Discern: Discovering counterfactual explanations using relevance features from neighbourhoods. In *2021 IEEE 33rd International Conference on Tools with Artificial Intelligence (ICTAI)*, pages 1466–1473. IEEE, 2021.

SF_PSY [Yang, 2022] Yujing Yang. On the study of chinese double-false counterfactual conditionals. In *2021 International Conference on Social Development and Media Commu- nication (SDMC 2021)*, pages 1553–1563. Atlantis Press, 2022.

SF_AI [Ye *et al.*, 2020] Xiaomeng Ye, David Leake, William Huibregtse, and Mehmet Dalkilic. Applying classto-class siamese networks to explain classifications with supportive and contrastive cases. In *International Conference on Case-Based Reasoning*, pages 245–260. Springer, 2020.

*REL [Ye *et al.*, 2021] Xiaomeng Ye, Ziwei Zhao, David Leake, Xizi Wang, and David Crandall. Applying the case difference heuristic to learn adaptations from deep network features. *arXiv preprint arXiv:2107.07095*, 2021.

*REL [Yousefzadeh and O'Leary, 2019] Roozbeh Yousefzadeh and Dianne P O'Leary. Interpreting neural networks using flip points. *arXiv preprint arXiv:1903.08789*, 2019.

REL [Zheng *et al.*, 2019] Jianyang Zheng, Hexing Zhu, Fangfang Chang, and Yunlong Liu. An improved relief feature selection algorithm based on monte-carlo tree search. *Systems Science & Control Engineering*, 7(1):304–310, 2019.

*CF [Zhao et al., 2022] Ziwei Zhao, David Leake, Xiaomeng Ye, and David Crandall. Generating counterfactual images: Towards a c2c-vae approach. 2022.

*REL [Čyras et al., 2021] Kristijonas Čyras, Antonio Rago, Emanuele Albini, Pietro Baroni, and Francesca Toni. Argumentative xai: A survey. In Proceedings of the Thirtieth International Joint Conference on Artificial Intelligence, IJCAI-21, pages 4392–4399. International Joint Conferences on Artificial Intelligence Organization, 8 2021. Survey Track.